\documentclass[11pt,letterpaper]{article}
\usepackage{naaclhlt2016}
\usepackage{times}
\usepackage{helvet}
\usepackage{courier}
\usepackage{latexsym}
\usepackage{url}
\usepackage{graphicx}
\usepackage{multirow}
\usepackage{amsmath}

\naaclfinalcopy 
\def\naaclpaperid{***} 

\newcounter{notecounter}
\newcommand{\enotesoff}{\long\gdef\enote##1##2{}}
\newcommand{\enoteson}{\long\gdef\enote##1##2{{
\stepcounter{notecounter}
\large\bf
\hspace{1cm}\arabic{notecounter} $<<<$ ##1: ##2
$>>>$\hspace{1cm}}}}
\enoteson
\enotesoff

\def\dnrm#1{\mbox{$_{\hbox{\scriptsize #1}}$}}
\def\figref#1{Figure~\ref{fig:#1}}
\def\figlabel#1{\label{fig:#1}\label{p:#1}}

\def\tabref#1{Table~\ref{tab:#1}}
\def\tablabel#1{\label{tab:#1}\label{p:#1}}

\def\secref#1{Section~\ref{sec:#1}}
\def\seclabel#1{\label{sec:#1}\label{p:#1}}
\def\eqref#1{Eq.~\ref{eqn:#1}}

\begin{document}

  \title{Comparing Convolutional Neural Networks \\to Traditional Models for Slot Filling}

\author{Heike Adel \and Benjamin Roth \and Hinrich Sch\"{u}tze \\
       Center for Information and Language Processing (CIS) \\ 
       LMU Munich\\
       Oettingenstr. 67, 80538 Munich, Germany\\ {\tt heike@cis.lmu.de}}

\date{}
 
\maketitle
\begin{abstract}
We address relation classification in the context of slot filling, the task of
finding and evaluating fillers like ``Steve Jobs'' for
the slot X in ``X founded Apple''. We propose a convolutional neural
network which splits the input sentence into three parts according to the 
relation arguments and compare it to state-of-the-art and traditional approaches
of relation classification. Finally, we combine
different methods and show that the combination is better 
than individual approaches. We also analyze the effect of genre 
differences on performance.
\end{abstract}

\section{Introduction}
Structured knowledge about the world is useful for many
natural language processing (NLP) tasks, such as
disambiguation, question answering or semantic search.
However, the extraction of structured information from
natural language text is challenging because one relation
can be expressed in many different ways. The TAC Slot
Filling (SF) Shared Task defines slot filling as extracting
fillers for a set of predefined relations (``slots'') from a
large corpus of text data. 
Exemplary relations are the city of birth of a person or
the employees or founders of a company.
Participants are provided
with an evaluation corpus and a query file consisting of pairs of entities
and slots. For each entity slot pair (e.g. ``Apple'' and ``founded\_by''), the systems 
have to return the second argument (``filler'') of the relation (e.g. ``Steve Jobs'')
as well as a supporting sentence from the evaluation corpus.
The key challenge in slot
filling is relation classification: given a sentence $s$ of the evaluation
corpus containing the name of a queried entity 
(e.g., ``Apple'') and a filler candidate (e.g., ``Steve Jobs''), we need
to decide whether $s$ expresses 
the relation (``founded\_by'', in this case).
We will refer to the mentions
of the two arguments of the relation as \emph{name} and
\emph{filler}.  Performance on relation classification
is crucial  for  slot filling since its effectiveness
directly depends on it.

In this paper, we investigate three complementary approaches
to relation classification.

The first approach is pattern matching, a
leading approach in the TAC evaluations. Fillers are
validated based on patterns.
In this work, we consider patterns learned with
distant supervision and patterns extracted from 
Universal Schema relations.

The second approach is support vector machines. 
We evaluate two different feature sets: a
bag-of-word feature set (BOW) and 
more sophisticated skip n-gram features.

Our third approach is a convolutional neural network (CNN).
CNNs have been applied to NLP tasks like sentiment analysis,
part-of-speech tagging and semantic role labeling. 
They can recognize phrase patterns
independent of their position in the sentence. Furthermore,
they make use of word embeddings that directly reflect
word similarity~\cite{mikolov}.  Hence, we expect them to be
robust models for the task of
classifying filler candidates and to generalize well
to unseen test data.
In this work, we train different variants of CNNs: 
As a baseline, we reimplement the recently developed
piecewise CNN 
\cite{piecewise}. Then, we extend this model
by splitting the contexts not only for pooling but
also for convolution (contextwise CNN).

Currently, there is no benchmark for slot filling.
Therefore, it is not possible to directly compare results that
were submitted to the Shared Task to new results. 
Comparable manual annotations for new results, for instance, cannot
be easily obtained.  
There are also many different system components,
such as document retrieval from the evaluation corpus
and coreference resolution, that
affect Shared Task performance and that are quite different
in nature from relation classification.
Even in the subtask of relation classification, it is
not possible to directly use existing relation classification 
benchmarks (e.g. \newcite{riedel}, \newcite{semevalData}) since 
data and relations can be quite different. 
Many 
benchmark relations, for instance, correspond to Freebase relations
but not all slots are modeled in Freebase and some
slots even comprise more than one Freebase relation.
While most relation classification benchmarks either
use newswire or web data, the SF task includes
documents from both domains (and discussion fora).
Another difference to traditional relation classification benchmarks
arises from the pipeline aspect of slot filling. Depending on 
the previous steps, the input for the relation
classification models can be incomplete, noisy,
include coreferent mentions, etc.

The official SF Shared Task evaluations only assess whole systems
(with potential subsequent faults in their pipelines \cite{pink}).
Thus, we expect component wise comparisons to be 
a valuable addition to the
Shared Task: With comparisons of single components,
teams would be able to improve their modules more specifically.
To start with one of the most important components, we have created a
benchmark for slot filling relation classification, 
based on 2012 -- 2014 TAC
Shared Task data. It will be described below and
published along with this paper.\footnote{\url{http://cistern.cis.lmu.de}} 
In addition to presenting model results on
this benchmark dataset, we also show that these results
correlate with end-to-end SF results. Hence, optimizing
a model on this dataset will also help improving results
in the end-to-end setting.

In our experiments, we found that our models suffer
from large genre differences in the TAC data.
Hence, the SF
Shared Task is a task that conflates an
investigation of domain (or genre) adaptation with the one of
slot filling. We argue that both problems are important NLP
problems and provide datasets and results for both
within and 
across genres. 
We hope that this new resource will encourage others
to  test their models on our dataset and that this will
help promote research on slot filling.

In summary, our contributions are as follows.
(i) We investigate 
the complementary strengths and weaknesses 
of different approaches to relation classification and 
show that their combination
can better deal with a diverse set of problems that slot filling
poses than each of the approaches individually.
(ii) We propose to split the context at the 
relation arguments before passing it to the CNN in order
to better deal with the special characteristics of a
sentence in relation classification. This outperforms
the state-of-the-art piecewise CNN.
(iii) We analyze the effect of genre on slot filling and
show that it is an important conflating variable that needs
to be carefully examined in research on slot filling.
(iv) We provide a benchmark for slot filling relation classification that 
will facilitate direct comparisons of 
models
in the future and show that results on this dataset 
are correlated with end-to-end system results.

\secref{relwork} gives an overview of related work.
\secref{challenges} discusses the challenges that slot
filling systems face.
In \secref{models}, we describe our slot filling models.
\secref{exResults} presents experimental setup and results.
\secref{analysis} analyzes the results.
We present our conclusions in 
\secref{conclusion}
and describe the resources we publish in 
\secref{resources}.

\section{Related Work}
\seclabel{relwork}
\textbf{Slot filling.}
The participants of the SF Shared Task \cite{sfOverview2013}
 are provided with a large text corpus. 
For evaluation, they get
a collection of queries and need to provide fillers for predefined relations
and an offset of a context which can serve as a justification. 
Most participants apply pipeline based systems.
\newcite{pink} analyzed sources of recall losses
in these pipelines.
The results
of the systems show the difficulty of the task: In the 2014
evaluation, the top-ranked system had an $F_1$ of .37~\cite{2014top}.
To train their models, most groups use distant
supervision~\cite{distant}. The top-ranked systems apply
machine learning based approaches rather than manually
developed patterns or models~\cite{sfOverview2014}.
The methods for extracting and scoring candidates range
from pattern based approaches~\cite{TALP,Sweat,PRIS,GDUFS,uschemaPat} over rule based 
systems~\cite{IIIT} to classifiers~\cite{sfCNN2012,roth2013}.
The top ranked system from 2013 used SVMs
and patterns for evaluating filler 
candidates~\cite{roth2013}; their results suggest that n-gram based features
are sufficient to build reliable classifiers for
the relation classification module.
They also show that SVMs outperform patterns.

\textbf{CNNs for relation classification.}
\newcite{zeng2014} and \newcite{dosSantos2015}
apply CNNs to the
relation classification SemEval Shared Task data from 2010
and show that CNNs  outperform other models.
We  train  CNNs on 
noisy distant supervised data since 
(in contrast to the SemEval Shared Task)
clean training sets are
not available.
\newcite{sfCNN2012} 
describe
a CNN for slot filling 
that is based on the output of a parser. 
We plan to explore parsing for creating a more linguistically motivated input
representation in the future.

\textbf{Baseline models.}
In this paper, we will compare our methods against traditional relation classification
models: Mintz++ \cite{distant,mimlre} and MIMLRE \cite{mimlre}.
Mintz++ is a model based on the Mintz features (lexical and syntactic
features for relation extraction). It was developed by \newcite{mimlre}
and used as a baseline model by them.
MIMLRE is a graphical model designed to cope with multiple instances and 
multiple labels in distant supervised data. It is trained with Expectation Maximization.

Another baseline model which we use in this work is
a piecewise convolutional neural network~\cite{piecewise}.
This recently published network
is designed especially for the relation classification task which
allows to split the context into three parts around the two relation arguments.
While it uses the whole context for convolution, it performs max pooling
over the three parts individually. In contrast, we propose
to split the context even earlier and apply the convolutional filters
to each part separately.

\textbf{Genre dependency.}
There are many studies showing the genre dependency of machine learning models.
In 2012, the SANCL Shared Task focused on 
evaluating models on web data that have been trained on news data~\cite{sancl}.
The results show that POS tagging performance can decline a
lot when the genre is changed.
For other NLP tasks like machine translation or sentiment analysis, this is also
a well-known challenge and domain adaptation has been
extensively studied~\cite{da1,da2}.
We do not investigate domain adaptation per se, but show that
the genre composition of the slot filling source corpus poses challenges
to genre independent models.

\section{Challenges of Slot Filling}
\seclabel{challenges}
Slot filling includes NLP challenges of various natures.
Given a large evaluation corpus, systems first need to
find documents relevant to the entity of the query.
This involves challenges like alternate names for the 
same entity, misspellings of names and ambiguous names 
(different entities with the same name).
Then for each relevant document, 
sentences with mentions of the entity need to be extracted,
as well as possible
fillers for the given slot. In most cases, 
coreference resolution
and named entity recognition
tools are used for these tasks.
Finally, the systems need to decide which 
filler candidate to output as the solution for the given slot.
This step can be reduced to relation classification.
It is one of the most crucial parts of the whole
pipeline since it directly influences the quality of the
final output.
The most important challenges for relation classification
for slot filling are little or noisy (distant supervised) training data,
data from different domains and
test sentences which have been extracted with a pipeline
of different NLP components. Thus, their
quality directly depends on the performance of the 
whole pipeline. If, for example, sentence splitting fails,
the input can be incomplete or too long. If coreference
resolution or named entity recognition fails, the 
relation arguments can be wrong or incomplete.

\section{Models for Relation Classification}
\seclabel{models}

\textbf{Patterns.}
The first approach we evaluate for relation classification
is pattern matching. 
For a given sentence, the pattern matcher classifies the relation
as correct if one of the patterns matches; otherwise the candidate is rejected.
In particular, we apply two different
pattern sets: The first set consists of patterns learned using
distant supervision (\textit{PATdist}). They have been used in the SF challenge
by the top-ranked system in the 2013 Shared Task~\cite{roth2013}. 
The second set contains patterns from universal schema relations
for the SF task (\textit{PATuschema}). Universal schema relations
are extracted based on matrix
factorization~\cite{riedel}.
In this work, we apply the universal schema patterns
extracted for slot filling by \newcite{uschemaPat}.

\textbf{Support vector machines (SVMs).}
Our second approach is support vector machines.
We evaluate two different feature sets:
bag-of-word features (\textit{SVMbow}) and skip n-gram features (\textit{SVMskip}).
Based on the results of \newcite{roth2013}, we will not use
additional syntactic or semantic features for our classifiers.
For SVMbow, the representation of a sentence consists 
of a flag and four bag-of-word
vectors.  Let $m_1$ and $m_2$ be the mentions of name and
filler (or filler and name) in the sentence, with $m_1$
occurring before $m_2$.  The binary flag indicates in which
order name and filler occur.
The four BOW vectors contain the words in the sentence
to the left of $m_1$, between $m_1$ and $m_2$, to the right
of $m_2$ and all words of the sentence. 
For SVMskip, we use the previously described
BOW features and additionally a feature vector which
contains skip n-gram features. They wildcard tokens in the middle
of the n-gram (cf.~\newcite{roth2013}). 
In particular, we use skip 3-grams, skip 4-grams
and skip 5-grams.
A possible skip 4-gram of the context ``, founder and director of'',
for example, would be the string ``founder of'', a 
pattern that could not have been directly extracted from this context
otherwise.
We train one linear SVM \cite{liblinear} for each relation
and feature set and tune parameter $C$ on dev.

\textbf{Convolutional neural networks (CNNs).}
CNNs are increasingly applied in NLP~\cite{cw,kalchbr}. 
They extract n-gram
based features independent of the position in
the sentence and
create (sub-)sentence representations. 
The two most important aspects that
make this possible are convolution and pooling.
Max pooling \cite{cw}
detects the globally most relevant features obtained by
local convolution.

Another promising aspect of CNNs for relation classification is that
they use an embedding based input representation.
With word embeddings, similar words are represented by
similar vectors and, thus, we can
recognize (near-)synonyms -- synonyms of relation triggers as well
as of other important context words.
If the CNN has learned, for example, that the context
``is based in'' triggers the relation
location\_of\_headquarters and that ``based'' has a
similar vector representation as ``located'', it may
recognize the context ``is located in'' correctly as another
trigger for the same relation even if it has never seen it
during training.
In the following paragraphs, we describe the different
variants of CNNs which we evaluate in this paper. For each
variant, we train one binary CNN per slot and optimize the 
number of filters ($\in \left\{300,1000,3000\right\}$), 
the size of the hidden layer ($\in \left\{100,300,1000\right\}$) 
and the filter width ($\in \left\{3,5\right\}$)
on dev.
We use word2vec
\cite{mikolov} to pre-train
word embeddings (dimensionality $d=50$) 
on a May-2014 English Wikipedia corpus.

\textit{Piecewise CNN.}
Our baseline CNN is the model developed by \newcite{piecewise}.
It represents the input sentence by a matrix of word vectors,
applies several filters for convolution and then
divides the resulting n-gram representation into left, middle and
right context based on the positions $m_1$
and $m_2$ of name and filler (see SVM description).
For each of the three parts, one max value is extracted
by pooling. The results are passed to a 
softmax classifier.

\begin{figure}
 \centering
 \includegraphics[width=\columnwidth]{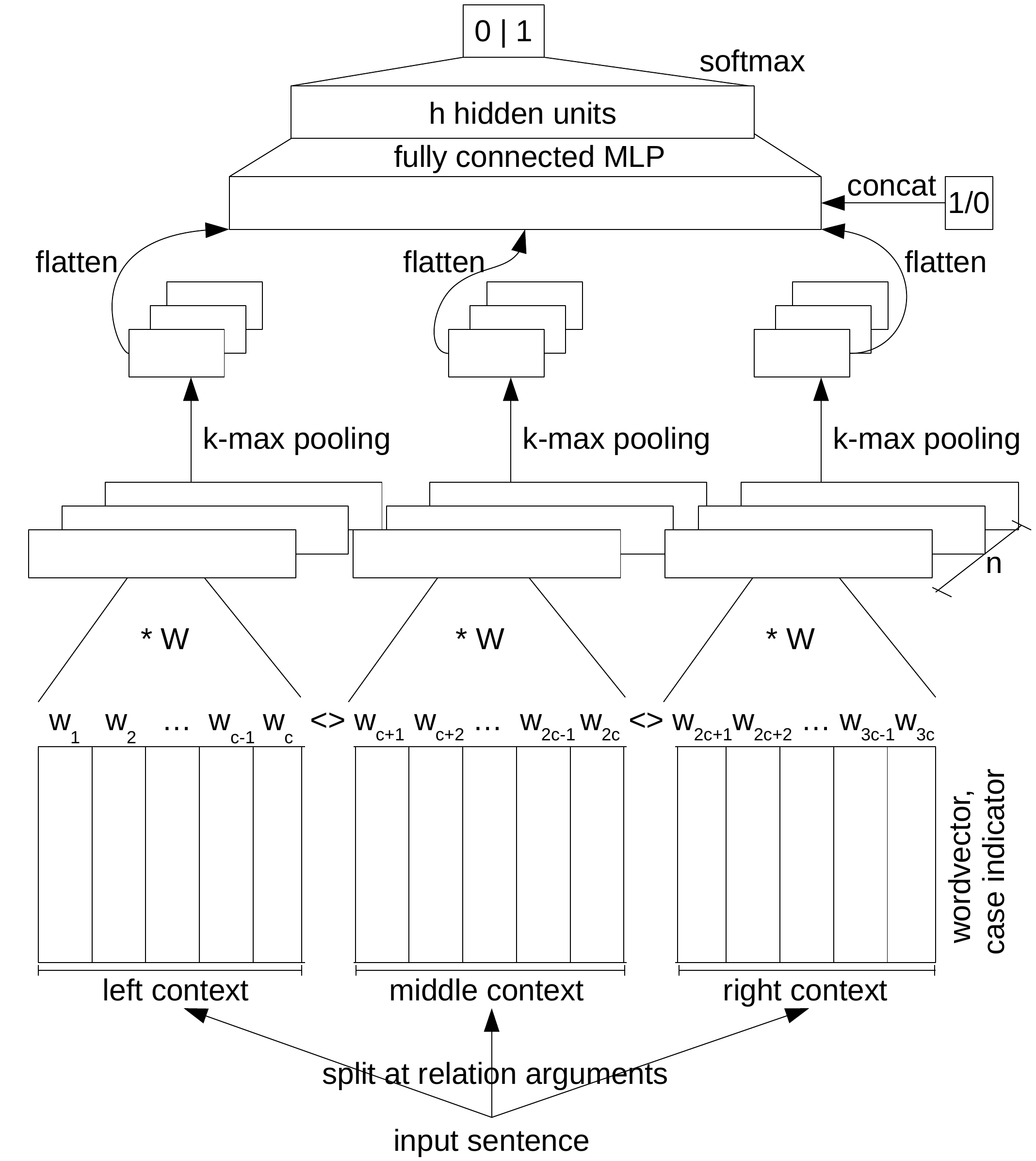}
 \caption{Contextwise CNN for relation classification}
 \figlabel{CNN}
\end{figure}

\textit{Contextwise CNN.}
In contrast to the piecewise CNN, we propose to split the context
before convolution as shown in \figref{CNN}. Hence, similar to our BOW vectors for the SVM, we
split the original context words into left, middle and right context.
Then, we apply convolution and pooling to each of the contexts separately.
In contrast to the piecewise CNN, there is no convolution across
relation arguments. Thus, the network learns to focus
on the context words and cannot be distracted
by the presence of (always present) relation arguments.
The filter weights $W$ are shared for the three contexts.
Our intuition is that the most important sequence features we
want to extract by convolution can appear in two or three of
the regions.
Weight sharing also 
reduces the number of
parameters and increases robustness. We also found in
initial experiments that
sharing  filter weights across left, middle, right
outperformed not sharing weights.
The results of convolution are pooled using $k$-max pooling~\cite{kalchbr}:
only the $k=3$ maximum values of each filter application are 
kept.
The pooling results are then concatenated to a single vector and extended by a flag
indicating whether the name or the filler appeared first in the sentence.

In initial experiments, we found that a
fully connected hidden layer after convolution and pooling
leads to a more powerful model. It connects the 
representations of the three contexts and,
thus, can draw conclusions based on cooccurring
 patterns across contexts.
Therefore, the result vector after convolution and pooling 
is fed into a fully connected hidden layer.
A softmax layer makes
the final decision.

For a fair comparison of models, we also
add a hidden layer to the piecewise CNN and apply $k$-max pooling there as well.
Thus, the number of parameters to learn is the same for both models.
We call this model \textit{CNNpieceExt}.
The key difference between CNNpieceExt and CNNcontext
is the time when the context is split into three parts: before or after
convolution. This affects the windows of words to which the convolutional
filters are applied.

\textbf{Model combination (CMB).}
\label{integration}
To combine a set $M$ of models for classification, we perform a 
simple linear combination of the scores of
the models:
\[
q\dnrm{CMB} = \sum_{m=1\ldots M}\alpha_m q_m
\]
where $q_m$ is the score of model $m$
and $\alpha_m$ is its weight (optimized on dev
using grid search). All weights sum to 1.

For a comparison of different combination
possibilities, see, for example, \cite{stacking}.

\section{Experiments and Results}
\seclabel{exResults}

\subsection{Training Data}
We used distant supervision for
generating training data. 
We created a set of (subject, relation, object) tuples by querying
Freebase \cite{freebase} 
for relations that correspond to the
slot relations. Then we scanned the following
corpora for sentences containing both arguments of a relation in the
tuple set:
(i) the TAC
source corpus \cite{sfTask}, 
(ii) a snapshot of Wikipedia (May 2014),
(iii) the Freebase description fields,
(iv) a subset of Clueweb\footnote{\url{http://lemurproject.org/clueweb12}},
(v) a New York Times corpus (LDC2008T19).
The resulting sentences are
positive training examples. 
Based on the tuple set, we selected negative examples 
by scanning the corpora for sentences that (i)
contain a mention of a name occurring in a tuple, 
(ii) do not contain the correct filler,
(iii) contain a mention different from the correct
filler, but with the same named entity type (based on
CoreNLP NER \cite{coreNLP}).  All negative examples for date slots,
for instance, are sentences containing an incorrect date.

This procedure gave us a large but noisy training set for most slots.
In order to reduce incorrect labels, we
applied a self-training procedure: We trained 
SVMs on the SF dataset created by \newcite{activeLearning}.
With the resulting SVMs, we predicted labels for
our training set. If the predicted label did not match the distant
supervised label, we deleted the corresponding training example \cite{nyt2012}.
This procedure was conducted in several iterations on different chunks
of the training set. Finally, the SF dataset and 
the filtered training examples were merged. (We do not
use the SF dataset directly because (i) it 
provides few examples per slot (min: 1, max: 4960) and
(ii) it consists of examples for which the classifiers of \newcite{activeLearning}
were indecisive, i.e., presumably contexts that are hard to classify.)
Since their contexts are similar, we also
merged city, state-or-province and country slots
to one location slot.

\begin{table*}[!t]
\small\centering
\begin{tabular}{l|rr|rr|rr|rr|rr|rr}
& \multicolumn{2}{c|}{Mintz++} & \multicolumn{2}{c|}{MIMLRE} & \multicolumn{2}{c|}{PATdist} & \multicolumn{2}{c|}{SVMskip} & \multicolumn{2}{c|}{CNNcontext} & \multicolumn{2}{c}{CMB} \\
 & dev & eval & dev & eval & dev & eval & dev & eval & dev & eval & dev & eval\\
\hline
per:age & .84 & .71 & .83 & .73 & .69 & \textbf{.80} & \textbf{.86} & .74 & .83 & .76 & \textbf{.86} & .77\\
per:alternate\_names & .29 & .03 & .29 & .03 & \textbf{.50} & \textbf{.50} & .35 & .02 & .32 & .04 & \textbf{.50} & \textbf{.50}\\
per:children & .76 & .43 & .77 & .48 & .10 & .07 & .81 & .68 & .82 & .61 & \textbf{.87} & \textbf{.76}\\
per:cause\_of\_death & .76 & .42 & .75 & .36 & .44 & .11 & \textbf{.82} & .32 & .77 & \textbf{.52} & \textbf{.82} & .31\\
per:date\_of\_birth & \textbf{1.0} & .60 & .99 & .60 & .67 & .57 & \textbf{1.0} & .67 & \textbf{1.0} & \textbf{.77} & \textbf{1.0} & .67\\
per:date\_of\_death & .67 & .45 & .67 & .45 & .30 & .32 & \textbf{.79} & \textbf{.54} & .72 & .48 & \textbf{.79} & \textbf{.54}\\
per:empl\_memb\_of & .38 & .36 & .41 & .37 & .24 & .22 & .42 & .36 & .41 & .37 & \textbf{.47} & \textbf{.39}\\
per:location\_of\_birth & .56 & .22 & .56 & .22 & .30 & .30 & .59 & .27 & .59 & .23 & \textbf{.74} & \textbf{.36}\\
per:loc\_of\_death & .65 & .41 & .66 & \textbf{.43} & .13 & .00 & .64 & .34 & .63 & .28 & \textbf{.70} & .35\\
per:loc\_of\_residence & .14 & .11 & .15 & .18 & .10 & .03 & \textbf{.31} & \textbf{.33} & .20 & .23 & \textbf{.31} & .31\\
per:origin & .40 & .48 & .42 & .46 & .13 & .11 & \textbf{.65} & \textbf{.64} & .43 & .39 & \textbf{.65} & .59\\
per:parents & .64 & .59 & .68 & .65 & .27 & .38 & .65 & \textbf{.79} & .65 & .78 & \textbf{.72} & .71\\
per:schools\_att & .75 & \textbf{.78} & .76 & .75 & .27 & .26 & .78 & .71 & .72 & .55 & \textbf{.79} & .71\\
per:siblings & \textbf{.66} & .59 & .64 & .59 & .14 & .50 & .60 & .68 & .63 & \textbf{.70} & .65 & \textbf{.70}\\
per:spouse & .58 & .23 & .59 & .27 & .40 & .53 & .67 & .32 & .67 & .30 & \textbf{.78} & \textbf{.57}\\
per:title & .49 & .39 & .49 & .40 & .48 & .42 & .54 & \textbf{.48} & .57 & .46 & \textbf{.59} & .46\\
org:alternate\_names & .49 & .46 & .50 & .48 & .70 & \textbf{.71} & .62 & .62 & .65 & .66 & \textbf{.72} & .67\\
org:date\_founded & .41 & .71 & .42 & \textbf{.73} & .47 & .40 & .57 & .70 & .64 & .71 & \textbf{.68} & .68\\
org:founded\_by & .60 & .62 & .70 & .65 & .39 & .62 & .77 & .74 & .80 & .68 & \textbf{.85} & \textbf{.77}\\
org:loc\_of\_headqu & .13 & .19 & .14 & .20 & .39 & .30 & .43 & .42 & .43 & .45 & \textbf{.50} & \textbf{.46}\\
org:members & .58 & .06 & .55 & .16 & .03 & \textbf{.29} & .70 & .13 & .65 & .04 & \textbf{.76} & .13\\
org:parents & .32 & .14 & .36 & .17 & .31 & .18 & .37 & .20 & .41 & .16 & \textbf{.52} & \textbf{.21}\\
org:subsidiaries & .32 & .43 & .35 & .35 & .32 & \textbf{.56} & .38 & .37 & .36 & .44 & \textbf{.42} & .49\\
org:top\_memb\_empl & .35 & .44 & .37 & .46 & .53 & .46 & .43 & \textbf{.55} & .43 & .53 & \textbf{.58} & .51\\
\hline
average & .53 & .41 & .54 & .42 & .35 & .36 & .62 & .48 & .60 & .46 & \textbf{.68} & \textbf{.53}\\
\end{tabular}
\caption{Performance on Slot Filling benchmark dataset (dev: data from 2012/2013, eval: from 2014). CMB denotes the
combination of PATdist, SVMskip and CNNcontext.}
\tablabel{yearwiseResults}
\end{table*}

\subsection{Evaluation Data}
\label{assessmentData}
One of the main challenges in building and evaluating relation classification
models for SF is the shortage of training and evaluation data.
Each group has their own datasets and comparisons
across groups are difficult.
Therefore, we have developed a script that creates a clean
dataset based on
manually annotated system outputs from 
previous Shared Task evaluations. In the future, it can be used
by all participants to evaluate components of their 
slot filling systems.\footnote{\url{http://cistern.cis.lmu.de}.
We publish scripts
since we cannot distribute
data.}
The script 
only extracts sentences that contain mentions of both
name and filler.
It conducts a heuristic check based on NER tags
to determine whether the name in the sentence is a
valid mention of the query name or is
referring to another entity.
In the latter case, the example is filtered out.
One difficulty is 
that many published offsets are
incorrect. We tried to match these using heuristics.
In general, we apply
filters that ensure high quality of the resulting
evaluation data even if that means that a considerable part of the TAC
system output is discarded.
In total, we extracted 39,386 high-quality evaluation instances 
out of the 59,755 system
output instances published by TAC and annotated as either completely correct or
completely incorrect.

A table in the supplementary material\footnote{also available at \url{http://cistern.cis.lmu.de}} gives statistics:
the number of positive and negative examples per slot and year (without duplicates).
For 2013, the most
examples were extracted. The lower number for 2014 is
probably due to
the newly introduced inference across
documents. This limits the number of sentences with
mentions of both
name and filler.  The average
ratio of positive to negative examples is 1:4. The number of
positive examples per slot and year ranges from 0 
(org:member\_of, 2014) to 581 (per:title, 2013),
the number of negative examples
from 5 (org:website, 2014) to 1886 (per:title, 2013).

In contrast to other relation classification benchmarks,
this dataset is not based on a knowledge base (such as Freebase)
and unrelated text (such as web documents)
but directly on the SF assessments. Thus, it
includes exactly the SF relations and 
addresses the challenges of the end-to-end task:
noisy data, possibly incomplete extractions of sentences
and data from different domains.

We use the data from 2012/2013 as development 
and the data from 2014 as evaluation set.

\subsection{Experiments}
\label{experiments}

We evaluate the models described in \secref{models},
select the best models  and combine them.

\textbf{Experiments with patterns.}
First, we compare the performance of PATdist and PATuschema on
our dataset. We evaluate the pattern matchers on all slots
presented in \tabref{yearwiseResults} and calculate their
average $F_1$ scores on dev. 
PATdist achieves a score of .35, PATuschema of .33.
Since it performs better, we
use PATdist in the following experiments.

\textbf{Experiments with SVMs.}
Second, we train and evaluate SVMbow and SVMskip.
Average $F_1$ of SVMskip 
and 
SVMbow are .62 and .59, respectively.
Thus, we use SVMskip.
We expected that SVMskip beats SVMbow due to its richer
feature set, but SVMbow performs
surprisingly well.

\textbf{Experiments with CNNs.}
Finally, we compare the performance of CNNpiece, CNNpieceExt and
CNNcontext.
While the baseline network CNNpiece~\cite{piecewise} achieves 
$F_1$ of .52 on dev, CNNpieceExt has an
$F_1$ score of .55 and
CNNcontext an $F_1$ of .60.
The difference of CNNpiece and CNNpieceExt
is due to the additional hidden layer
and k-max pooling.
The considerable difference in performance
of CNNpieceExt and CNNcontext
shows that splitting the context for convolution
has a positive effect on the performance of the network.

\textbf{Overall results.}
\tabref{yearwiseResults} shows the slot wise
results of the best patterns (PATdist), SVMs 
(SVMskip) and CNNs (CNNcontext).
Furthermore, it provides a comparison with 
two baseline models: Mintz++ and MIMLRE.
SVM and CNN clearly outperform these baselines.
They also outperform PAT for almost all slots.
The difference between dev and eval results varies a lot
among the slots. We suspect that this is a result
of genre differences in the data and analyze this
in \secref{genretime}.

Slot wise results of the other models (PATuschema, SVMbow,
CNNpiece, CNNpieceExt) can be found in the supplementary 
material.


Comparing PAT, SVM and CNN,\footnote{In prior experiments, we also 
compared with recurrent neural networks.
 RNN
performance was comparable to  CNNs, but required
much more training time and parameter tuning. Therefore, we
focus on CNNs in this paper.
See also
\newcite{vu16rnncnn}.} 
different patterns emerge
for different slots. Each is best on a subset of the slots
(see bold numbers). This indicates that relation classification
for slot filling is not
a uniform problem: each slot has special
properties and the three approaches are good at modeling a
different subset of these properties. Given the big differences,
we expect to gain
performance  by combining the
three approaches. 
Indeed, CMB (PATdist + SVMskip + CNNcontext), the
combination of the three best performing models,
obtains the best results in average (in bold). 

\secref{correlation}
shows that the performance on our dataset is highly
correlated with SF end-to-end performance.
Thus, our results indicate that a combination of different
models ist the most promising approach to getting good
performance on slot filling. 

\section{Analysis}
\seclabel{analysis}
\subsection{Contribution of Each Model}
\begin{figure}
\centering
\includegraphics[width=\columnwidth]{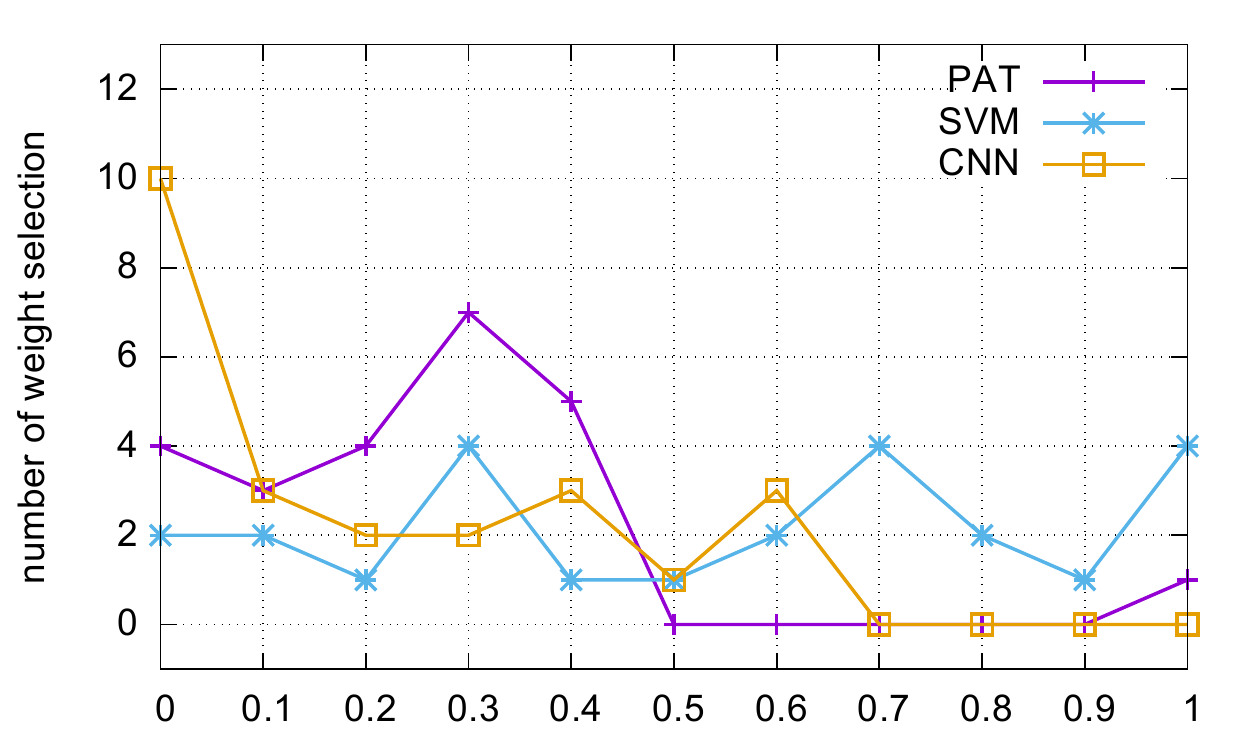}
\caption{\#  times each weight is selected in CMB}
\label{interpolationWeights}
\end{figure}

To see how much each model contributes to CMB, we count
how often each weight between 0.0 and 1.0 is selected
for the linear interpolation.
The results are plotted
as a histogram (Figure \ref{interpolationWeights}). 
A weight of 0.0 means that the corresponding model
does not contribute to CMB. We see
that all three models contribute to CMB for most of the slots. 
The CNN, for instance, is included in the combination for 14
of 24 slots.

\subsection{Comparison of CNN to Traditional Models}

Our motivation for using a CNN is
that convolution and max pooling can recognize important n-grams independent
of their position in the sentence. 
To investigate this effect,
we select for each CNN the top five kernels whose activations are 
the most correlated with the final score of the positive
class.
Then we calculate which n-grams are selected by these kernels
in the max pooling step. This corresponds to those n-grams which are recognized by the kernel to be the
most informative for the given slot.
\figref{CNNanalysis} shows the result for an example sentence 
expressing the slot relation org:parents.
The height of a bar is the number of times that the 3-gram around the corresponding word
was selected by $k$-max pooling; e.g., the bar above
``newest'' corresponds to the trigram ``its newest subsidiary''.
The figure shows that the convolutional filters are able to learn phrases that 
trigger a relation, e.g., ``its subsidiary''. In contrast to
patterns, they do not rely on exact matches.
The first reason is embeddings. They 
generalize  similar words and phrases by assigning
similar word vectors to them.
For PAT and SVM, this type of generalization is more difficult.
The second type of generalization that the CNN learns
concerns insertions, similar to skip n-gram features.
The recognition of important phrases 
in convolution 
is robust against insertions. An example is
``newest'' in \figref{CNNanalysis}, a word that is not
important for the slot.
\begin{figure}
\centering
\includegraphics[width=\columnwidth]{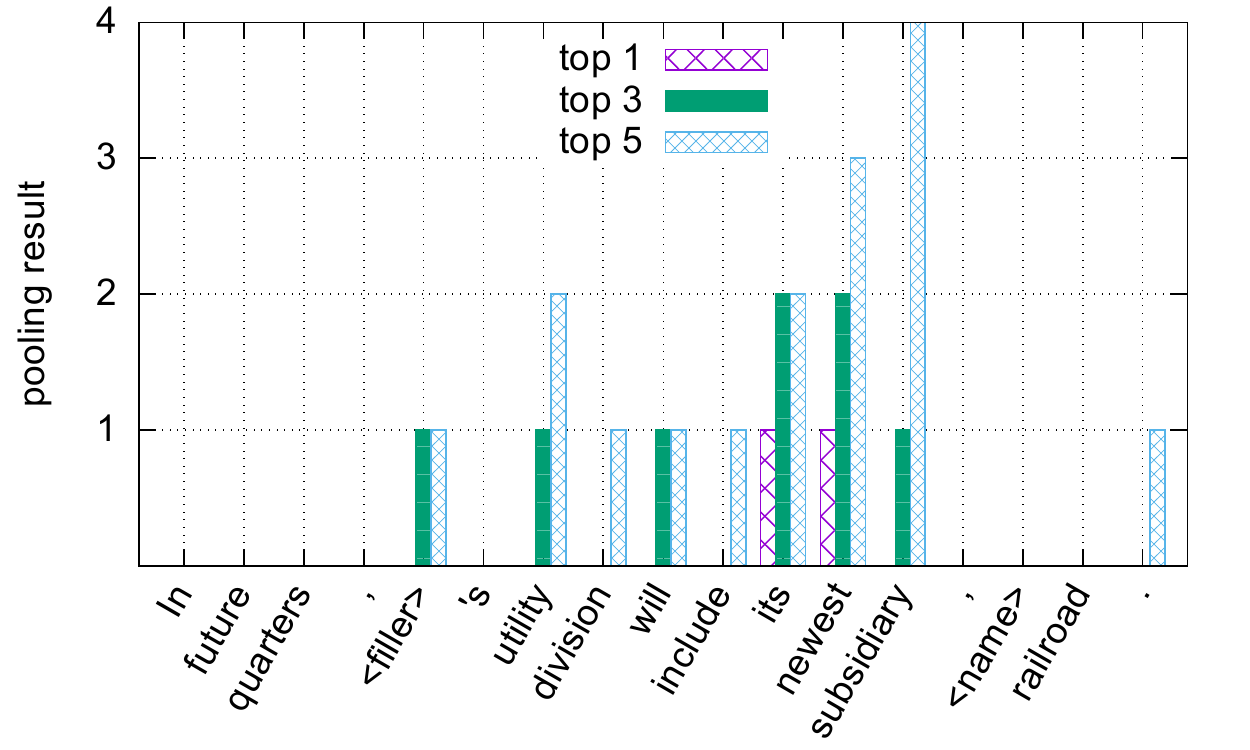}
\caption{Analysis of convolution and pooling}
\figlabel{CNNanalysis}
\end{figure}

A direct comparison of results with PAT shows that
the CNN has better eval scores for about
67\% of the slots (see \tabref{yearwiseResults}).
Our reasoning above can explain this.
Compared to the SVM, the CNN generalizes better to unseen
data in only 42\% of all cases. 
The fact that this does not happen in more cases
shows the
power of the skip n-gram features of the SVM: they also provide
a kind of generalization against insertions.
The SVM might also need less data to train than the CNN.
Nevertheless, the final scores show that the CNN
performs almost as well as the SVM in average (.60 vs .62 on dev,
.46 vs .48 on eval) and
contributes to a better combination score.

\subsection{Correlation with End-to-end Results}
\seclabel{correlation}
In this section, we show that using the dataset we provide with this
paper allows tuning classification models for 
the end-to-end SF task. For each model and each possible combination of models, we 
calculate average results on
our evaluation set as well as final $F_1$ scores when running the whole slot filling
pipeline with our in-house system. 
The best results of our slot filling system are an $F_1$ of .29
on the 2013 queries and of .25 on the 2014
queries. 
We calculate Pearson's correlation coefficient to assess
correlation of relation classification and end-to-end performances
for the $n$ different system configurations (i.e., model combinations).
The correlation of the results on our 
eval dataset with the SF results on 2013 queries is .89, the
correlation with the SF results on 2014 queries is .82.
This confirms that good results on the dataset we propose
lead to good results on the slot filling end-to-end task.

\subsection{Effect of Genre and Time}
\seclabel{genretime}
The TAC source corpus consists of about 1M
news documents, 1M web documents and 100K documents from discussion 
forums~\cite{sfTask}.
The distribution of these different genres
in the extracted assessment data
is as shown in \tabref{genredist}.

\begin{table}[t]
\centering
\small
 \begin{tabular}{l|r|r}
  & 2012/3 & 2014\\
  \hline
  news & 87.5\% & 73.4\% \\
  web + forums & 12.5\% & 26.6\%
 \end{tabular}
\caption{Distribution of genres\tablabel{genredist}}
\end{table}

The proportion of non-news more than doubled from
12.5\% to 26.6\%.  
Thus, when using 2012/2013 as
the development and 2014 as the test set, we are faced with
a domain adaptation problem.

In this section, we show the effect of
domain differences on our models in more detail.
For our genre analysis, we retrain our models on genre specific training sets 
WEB and NEWS$_{\subset}$
and 
show within-genre
as well as cross-genre evaluations.
To
avoid performance differences due to different training
set sizes, we reduced the news training set to the same size
as the web training set. We refer to this subset as
NEWS$_{\subset}$.

\textbf{Cross-genre evaluation.}
\tabref{crossgenre}
shows results of
testing models trained on genre-specific data: on data of the same genre
and on data of  the other genre. We present results only for a subset of relations
in this paper, however, the numbers for the other slots follow the same trends.

Models trained on news (left part) show clearly higher performance in the within-genre
evaluation than cross-genre. For models trained on web (right part), this is different.
We suspect that the reason is that web data is much noiser
and thus less predictable, even for models trained on web.
For all evaluations, the differences among dev and eval are quite large.
Especially for slot filling on web (bottom part of~\tabref{crossgenre}),
the results on dev do not seem much related to the results on eval.
This domain effect increases the difficulties of training robust relation classification
models for slot filling. It can also explain why optimizing models
for unseen data (with unknown genre distributions) as in \tabref{yearwiseResults} is challenging.
Since slot filling by itself is a challenging 
task, even in the absence of domain differences, we will distribute two splits: a split by
year and a split by genre.
For training and tuning models for the slot filling research challenge, the 
year split can be used to cover the challenge of mixing different genres.
For experiments on  domain adaptation or genre-specific effects,
our genre split can be used.

 \begin{table}[t]
  \centering
  \footnotesize
   \setlength{\tabcolsep}{3.5pt}
  \begin{tabular}{rl|rr|rr||rr|rr}
   & &\multicolumn{4}{c||}{Train on NEWS$_{\subset}$} & \multicolumn{4}{c}{Train on WEB} \\
  & & \multicolumn{2}{c|}{SVM} & \multicolumn{2}{c||}{CNN} & \multicolumn{2}{c|}{SVM} &  \multicolumn{2}{c}{CNN}\\
    & & dev & ev & dev & ev & dev & ev & dev & ev\\
   \hline\hline
   \multirow{6}{*}{\rotatebox{90}{Test on news}} &
 per:age & .79 & .80 & \textbf{.88} & \textbf{.87} & .78 & .76 & \textbf{.85} & \textbf{.83} \\
& per:children & \textbf{.85} & \textbf{.86} & .78 & .78 & \textbf{.75} & \textbf{.80} & .00 & .07 \\
& per:spouse & .74 & .64 & \textbf{.76} & \textbf{.71} & \textbf{.77} & .65 & .73 & \textbf{.67} \\
& org:alt\_names & .22 & .32 & \textbf{.69} & \textbf{.67} & .65 & \textbf{.70} & \textbf{.66} & .66 \\
& org:loc\_headqu & .51 & .50 & \textbf{.53} & \textbf{.51} & .51 & \textbf{.53} & \textbf{.53} & .50 \\
& org:parents & \textbf{.30} & .32 & .29 & \textbf{.34} & .26 & .33 & \textbf{.30} & \textbf{.34}\\
   \cline{2-10}
   \multirow{6}{*}{\rotatebox{90}{Test on web}} &
 per:age & .33 & .73 & \textbf{.57} & \textbf{.83} & .00 & .67 & \textbf{.57} & \textbf{.83} \\
&  per:children & .59 & \textbf{.33} & \textbf{.70} & \textbf{.33} & \textbf{.63} & \textbf{.57} & .00 & .00 \\
&  per:spouse & .52 & .50 & \textbf{.60} & \textbf{.57} &  .56 & .57 & \textbf{.67} & \textbf{.62} \\
&  org:alt\_names & .27 & .19 & \textbf{.51} & \textbf{.37} & \textbf{.60} & \textbf{.49} & .56 & .38 \\
&  org:loc\_headqu & .39 & \textbf{.46} & \textbf{.43} & .44 & \textbf{.44} & \textbf{.48} & .36 & .47\\
&  org:parents & .09 & \textbf{.08} & \textbf{.11} & .07  & .10 & \textbf{.08} & \textbf{.15} & \textbf{.08} \\
\end{tabular}
 \caption{Genre specific $F_1$ scores. Genre specific training data (of the same sizes).
  Top: news results. Bottom: web results.
  \tablabel{crossgenre}
  }
  \end{table}

\section{Conclusion}
\seclabel{conclusion}
In this paper, we presented different approaches to slot
filling relation classification: patterns, support vector machines
and convolutional neural networks.
We investigated their
complementary strengths and weaknesses and
showed that their combination
can better deal with a diverse set of problems that slot filling
poses than each of the approaches individually.
We proposed a contextwise CNN which outperforms the recent 
state-of-the-art piecewise CNN.
Furthermore, we analyzed the effect of genre on slot filling
and showed that it needs to be carefully examined in
research on slot filling.
Finally, we provided a benchmark for slot filling relation classification
that will facilitate direct comparisons of
approaches in the future.

\section{Additional Resources}
\seclabel{resources}
We publish the scripts that we developed
to extract the annotated evaluation data and 
our splits by genre and by year as well as the dev/eval splits.


\section*{Acknowledgments}
Heike Adel is a recipient of the Google European Doctoral
Fellowship in Natural Language Processing and this
research is supported by this fellowship.

This research was also supported by Deutsche
Forschungsgemeinschaft: grant SCHU 2246/4-2.

We would like to thank Gabor Angeli for his help
with the Mintz++ and MIMLRE models.

\newpage

\bibliographystyle{naaclhlt2016}

\bibliography{adel.bib}

\end{document}